\documentclass[11pt,a4paper]{article}
\usepackage[hyperref]{acl2018}
\usepackage{times}
\usepackage{amsmath}
\usepackage{latexsym}
\usepackage{booktabs}
\usepackage[nolist]{acronym}
\usepackage{url}
\usepackage{todonotes}
\usepackage{array}
\newcolumntype{L}{>{\arraybackslash}m{13cm}}

\usepackage[utf8]{inputenc}
\usepackage[russian,english]{babel}

\aclfinalcopy 
\def\aclpaperid{1108} 

\title{NeuralREG: An end-to-end approach to referring expression generation}

\author{Thiago Castro Ferreira$^{1}$ \, Diego Moussallem$^{2,3}$ \, \'{A}kos K\'{a}d\'ar$^{1}$ \, Sander Wubben$^{1}$ \, Emiel Krahmer$^{1}$ \\
         $^{1}$Tilburg center for Cognition and Communication (TiCC), Tilburg University, The Netherlands \\
         $^{2}$AKSW Research Group, University of Leipzig, Germany \\
         $^{3}$Data Science Group, University of Paderborn, Germany \\
         {\tt \{tcastrof,a.kadar,s.wubben,e.j.krahmer\}@tilburguniversity.edu} \\
         {\tt moussallem@informatik.uni-leipzig.de}}

\date{}
\begin{acronym}[UML]
	\acro{AOS}{Agricultural Ontology Services}
	\acro{AGRIS}{Agricultural Science and Technology}
	\acro{API}{Application Programming Interface}
	\acro{A2KB}{Annotation to Knowledge Base}
	\acro{BPSO}{Binary Particle-Swarm Optimization}
	\acro{BPMLOD}{Best Practices for Multilingual Linked Open Data}
	\acro{BFS}{Breadth-First-Search}
	\acro{CBD}{Concise Bounded Description}
	\acro{COG}{Content Oriented Guidelines}
	\acro{CSV}{Comma-Separated Values}
	\acro{CCR}{cross-document co-reference}
	\acro{DPSO}{Deterministic Particle-Swarm Optimization}
	\acro{DALY}{Disability Adjusted Life Year}
	\acro{D2KB}{Disambiguation to Knowledge Base}

	\acro{ER}{Entity Resolution}
	\acro{EM}{Expectation Maximization}
	\acro{EL}{Entity Linking}
	\acro{FAO}{Food and Agriculture Organization of the United Nations}
	\acro{GIS}{Geographic Information Systems}
	\acro{GHO}{Global Health Observatory}
	\acro{HDI}{Human Development Index}
	\acro{ICT}{Information and communication technologies}
    \acro{KB}{Knowledge Base}
	\acro{LR}  {Language Resource}
	\acro{LD}  {Linked Data}
	\acro{LLOD}  {Linguistic Linked Open Data}
	\acro{LIMES}{LInk discovery framework for MEtric Spaces}
	\acro{LS}  {Link Specifications}
	\acro{LDIF}{Linked Data Integration Framework}
	\acro{LGD} {LinkedGeoData}
	\acro{LOD} {Linked Open Data}
    \acro{LSTM} {Long-Short Term Memory Layers}
	\acro{MSE}{Mean Squared Error}
	\acro{MWE}{Multiword Expressions}
	\acro{NIF}{NLP Interchange Format}
	\acro{NIF4OGGD}{NLP Interchange Format for Open German Governmental Data}
	\acro{NLP}{Natural Language Processing}
	\acro{NER}{Named Entity Recognition}
	\acro{NED}{Named Entity Disambiguation}
	\acro{NEL}{Named Entity Linking}
	\acro{NN}{Neural Network}
    \acro{NLG}{Natural Language Generation}
	\acro{OSM}{OpenStreetMap}
	\acro{OWL}{Web Ontology Language}
	\acro{PFM}{Pseudo-F-Measures}
	\acro{PSO}{Particle-Swarm Optimization}
	\acro{QA}{Question Answering}
	\acro{RDF}{Resource Description Framework}
    \acro{REG}{Referring Expression Generation}
	\acro{SKOS}{Simple Knowledge Organization System}
	\acro{SPARQL}{SPARQL Protocol and RDF Query Language}
	\acro{SRL}{Statistical Relational Learning}
	\acro{SF}{surface forms}
    \acro{SW}{Semantic Web}
	\acro{UML}{Unified Modeling Language}
	\acro{WHO}{World Health Organization}
	\acro{WKT}{Well-Known Text}
	\acro{W3C}{World Wide Web Consortium}
	\acro{YPLL}{Years of Potential Life Lost}
\end{acronym}  
\begin{document}
\maketitle
\begin{abstract}
 Traditionally, Referring Expression Generation (REG) models first decide on the form and then on the content of references to discourse entities in text, typically relying on features such as salience and grammatical function. In this paper, we present a new approach (NeuralREG), relying on deep neural networks, which makes decisions about form and content in one go without explicit feature extraction. Using a delexicalized version of the WebNLG corpus, we show that the neural model substantially improves over two strong baselines. Data and models are publicly available\footnote{\url{https://github.com/ThiagoCF05/NeuralREG}}.
\end{abstract}

\section{Introduction}

\ac{NLG} is the task of automatically converting non-linguistic data into coherent natural language text \cite{reiter2000,gatt2017}. Since the input data will often consist of entities and the relations between them, generating references for these entities is a core task in many NLG systems \cite{dale1995,krahmer2012}. \ac{REG}, the task responsible for generating these references, is typically presented as a two-step procedure. First, the referential form needs to be decided, asking whether a reference at a given point in the text should assume the form of, for example, a proper name (``Frida Kahlo''), a pronoun (``she'') or description (``the Mexican painter''). In addition, the REG model must account for the different ways in which a particular referential form can be realized. For example, both ``Frida'' and ``Kahlo'' are name-variants that may occur in a text, and she can alternatively also be described as, say, ``the famous female painter''.

Most of the earlier \ac{REG} approaches focus either on selecting referential form \cite{orita2015,ferreira2016b}, or on selecting referential content, typically zooming in on one specific kind of reference such as a pronoun  \cite[e.g.,][]{henschel2000,callaway2002}, definite description \cite[e.g.,][]{dale1991,dale1995} or proper name generation \cite[e.g.,][]{siddharthan2011,deemter2016,ferreira2017}. Instead, in this paper, we propose NeuralREG: an end-to-end approach addressing the full REG task, which given a number of entities in a text, produces corresponding referring expressions, simultaneously selecting both form and content. Our approach is based on neural networks which generate referring expressions to discourse entities relying on the surrounding linguistic context, without the use of any feature extraction technique.

Besides its use in traditional pipeline NLG systems \cite{reiter2000}, \ac{REG} has also become relevant in modern ``end-to-end'' NLG approaches, which perform the task in a more integrated manner \cite[see e.g.][]{konstas2017,claire2017b}. Some of these approaches have recently focused on inputs which references to entities are delexicalized to general tags (e.g., ENTITY-1, ENTITY-2) in order to decrease data sparsity. Based on the delexicalized input, the model generates outputs which may be likened to templates in which references to the discourse entities are not realized (as in ``The ground of ENTITY-1 is located in ENTITY-2.'').

While our approach, dubbed as NeuralREG, is compatible with different applications of REG models, in this paper, we concentrate on the last one, relying on a specifically constructed set of 78,901 referring expressions to 1,501 entities in the context of the semantic web, derived from a (delexicalized) version of the WebNLG corpus \cite{claire2017,claire2017b}. Both this data set and the model will be made publicly available. We compare NeuralREG against two baselines in an automatic and human evaluation, showing that the integrated neural model is a marked improvement.

\section{Related work}

In recent years, we have seen a surge of interest in using (deep) neural networks for a wide range of NLG-related tasks, as the generation of (first sentences of) Wikipedia entries \cite[][]{lebret2016}, poetry \cite[][]{zhang2014}, and texts from abstract meaning representations \cite[e.g.,][]{konstas2017,ferreira2017b}. However, the usage of deep neural networks for REG has remained limited and we are not aware of any other integrated, end-to-end model for generating referring expressions in discourse.

\begin{figure*}
\footnotesize{
\begin{center}
\begin{tabular}{l l l}
\textbf{Subject} & \textbf{Predicate}        & \textbf{Object} \\
108\_St\_Georges\_Terrace & location       & Perth \\
Perth                     & country        & Australia \\
108\_St\_Georges\_Terrace & completionDate & 1988\textit{@year} \\
108\_St\_Georges\_Terrace & cost           & 120 million (Australian dollars)\textit{@USD} \\
108\_St\_Georges\_Terrace & floorCount     & 50\textit{@Integer} \\
\end{tabular}
\vspace{0.2cm}
\begin{center}
$\downarrow$
\vspace{0.2cm}

108 St Georges Terrace was completed in 1988 in Perth, Australia. It has a total of 50 floors and cost 120m Australian dollars.
\end{center}

\caption{Example of a set of triples (top) and corresponding text (bottom).}
\label{fig:triple}
\end{center}
}
\end{figure*}

There is, however, a lot of earlier work on selecting the form and content of referring expressions, both in psycholinguistics and in computational linguistics. In psycholinguistic models of reference, various linguistic factors have been proposed as influencing the form of referential expressions, including cognitive status \cite{gundel1993}, centering \cite{grosz1995} and information density \cite{jaeger2010}. In models such as these, notions like  \textit{salience} play a central role, where it is assumed that entities which are salient in the discourse are more likely to be referred to using shorter referring expressions (like a pronoun) than less salient entities, which are typically referred to using longer expressions (like full proper names).

Building on these ideas, many REG models for generating references in texts also strongly rely on the concept of salience and factors contributing to it. \citet{reiter2000} for instance, discussed a straightforward rule-based method based on this notion, stating that full proper names can be used for initial references, typically less salient than subsequent references, which, according to the study, can be realized by a pronoun in case there is no mention to any other entity of same person, gender and number between the reference and its antecedents. More recently, \citet{ferreira2016b} proposed a data-driven, non-deterministic model for generating referential forms, taking into account salience features extracted from the discourse such as grammatical position, givenness and recency of the reference. Importantly, these models do not specify which contents a particular reference, be it a proper name or description, should have. To this end, separate models are typically used, including, for example, \citet{dale1995} for generating descriptions, and \citet{siddharthan2011,deemter2016} for proper names.

Of course, when texts are generated in practical settings, both form and content need to be chosen. This was the case, for instance, in the GREC shared task \cite{belz2010}, which aimed to evaluate models for automatically generated referring expressions grounded in discourse. The input for the models were texts in which the referring expressions to the topic of the relevant Wikipedia entry were removed and appropriate references throughout the text needed to be generated (by selecting, for each gap, from a list of candidate referring expressions of different forms and with different contents). Some participating systems approached this with traditional pipelines for selecting referential form, followed by referential content, while others proposed more integrated methods. More details about the models can be seen on \citet{belz2010}.

In sum, existing REG models for text generation strongly rely on abstract features such as the salience of a referent for deciding on the form or content of a referent. Typically, these features are extracted automatically from the context, and engineering relevant ones can be complex. Moreover, many of these models only address part of the problem, either concentrating on the choice of referential form or on deciding on the contents of, for example, proper names or definite descriptions. In contrast, we introduce NeuralREG, an end-to-end approach based on neural networks which generates referring expressions to discourse entities directly from a delexicalized/wikified text fragment, without the use of any feature extraction technique. Below we describe our model in more detail, as well as the data on which we develop and evaluate it. 

\section{Data and processing}

\subsection{WebNLG corpus}

Our data is based on the WebNLG corpus \cite{claire2017}, which is a parallel resource initially released for the eponymous NLG challenge. In this challenge, participants had to automatically convert non-linguistic data from the Semantic Web into a textual format \cite{claire2017b}. The source side of the corpus are sets of \textit{Resource Description Framework} (RDF) triples. Each RDF triple is formed by a Subject, Predicate and Object, where the Subject and Object are constants or Wikipedia entities, and predicates represent a relation between these two elements in the triple. The target side contains English texts, obtained by \textit{crowdsourcing}, which describe the source triples. Figure \ref{fig:triple} depicts an example of a set of 5 RDF triples and the corresponding text.

The corpus consists of  25,298 texts describing 9,674 sets of up to 7 RDF triples (an average of 2.62 texts per set) in 15 domains \cite{claire2017b}. In order to be able to train and evaluate our models for referring expression generation (the topic of this study), we produced a delexicalized version of the original corpus.

\subsection{Delexicalized WebNLG}

\begin{figure*}
\footnotesize{
\begin{center}
\begin{tabular}{l l }
\textbf{Tag} & \textbf{Entity} \\
AGENT-1   & 108\_St\_Georges\_Terrace \\
BRIDGE-1  & Perth \\
PATIENT-1 & Australia \\
PATIENT-2 & 1988\textit{@year} \\
PATIENT-3 & ``120 million (Australian dollars)''\textit{@USD} \\
PATIENT-4 & 50\textit{@Integer} \\
\end{tabular}

\vspace{0.3cm}
\begin{center}
\textbf{AGENT-1} was completed in \textbf{PATIENT-2} in \textbf{BRIDGE-1} , \textbf{PATIENT-1} . \textbf{AGENT-1} has a total of \textbf{PATIENT-4} floors and cost \textbf{PATIENT-3} .

\vspace{0.2cm}
$\downarrow_{Wiki}$
\vspace{0.2cm}

\textbf{108\_St\_Georges\_Terrace} was completed in \textbf{1988} in \textbf{Perth} , \textbf{Australia} . \textbf{108\_St\_Georges\_Terrace} has a total of \textbf{50} floors and cost \textbf{20\_million\_(Australian\_dollars)} .
\end{center}

\caption{Mapping between tags and entities for the related delexicalized/wikified templates.}
\label{fig:delex}
\end{center}
}
\end{figure*}

We delexicalized the training and development parts of the WebNLG corpus by first automatically mapping each entity in the source representation to a general tag. All entities that appear on the left and right side of the triples were mapped to AGENTs and PATIENTs, respectively. Entities which appear on both sides in the relations of a set were represented as BRIDGEs. To distinguish different AGENTs, PATIENTs and BRIDGEs in a set, an  ID was given to each entity of each kind (PATIENT-1, PATIENT-2, etc.). Once all entities in the text were mapped to different roles, the first two authors of this study manually replaced the referring expressions in the original target texts by their respective tags. Figure \ref{fig:delex} shows the entity mapping and the delexicalized template for the example in Figure \ref{fig:triple} in its versions representing the references with general tags and Wikipedia IDs. 

We delexicalized 20,198 distinct texts describing 7,812 distinct sets of RDF triples, resulting in 16,628 distinct templates. While this dataset (which we make available) has various uses, we used it to extract a collection of referring expressions to Wikipedia entities in order to evaluate how well our REG model can produce references to entities throughout a (small) text.

\subsection{Referring expression collection}
\label{sec:reg_corpus}

Using the delexicalized version of the WebNLG corpus, we automatically extracted all referring expressions by tokenizing the original and delexicalized versions of the texts and then finding the non overlapping items. For instance, by processing the text in Figure \ref{fig:triple} and its delexicalized template in Figure \ref{fig:delex}, we would extract referring expressions like ``108 St Georges Terrace'' and ``It'' to \textit{$\langle$ AGENT-1, 108\_St\_Georges\_Terrace $\rangle$}, ``Perth'' to \textit{$\langle$ BRIDGE-1, Perth $\rangle$}, ``Australia'' to \textit{$\langle$ PATIENT-1, Australia $\rangle$} and so on.

Once all texts were processed and the referring expressions extracted, we filtered only the ones referring to Wikipedia entities, removing references to constants like dates and numbers, for which no references are generated by the model. In total, the final version of our dataset contains 78,901 referring expressions to 1,501 Wikipedia entities, in which 71.4\% (56,321) are proper names, 5.6\% (4,467) pronouns, 22.6\% (17,795) descriptions and 0.4\% (318) demonstrative referring expressions. We split this collection in training, developing and test sets, totaling 63,061, 7,097 and 8,743 referring expressions in each one of them.


Each instance of the final dataset consists of a truecased tokenized referring expression, the target entity (distinguished by its Wikipedia ID), and the discourse context preceding and following the relevant reference (we refer to these as the pre- and pos-context). Pre- and pos-contexts are the lowercased, tokenized and delexicalized pieces of text before and after the target reference. References to other discourse entities in the pre- and pos-contexts are represented by their Wikipedia ID, whereas constants (numbers, dates) are represented by a one-word ID removing quotes and replacing white spaces with underscores (e.g., \textit{120\_million\_(Australian\_dollars)} for ``120 million (Australian dollars)'' in Figure \ref{fig:delex}). 

Although the references to discourse entities are represented by general tags in a delexicalized template produced in the generation process (AGENT-1, BRIDGE-1, etc.), for the purpose of disambiguation, NeuralREG's inputs have the references represented by the Wikipedia ID of their entities. In this context, it is important to observe that the conversion of the general tags to the Wikipedia IDs can be done in constant time during the generation process, since their mapping, like the first representation in Figure \ref{fig:delex}, is the first step of the process. In the next section, we show in detail how NeuralREG models the problem of generating a referring expression to a discourse entity.

\section{NeuralREG}

NeuralREG aims to generate a referring expression $y = \lbrace y_{1}, y_{2}, ... , y_{T} \rbrace$ with $T$ tokens to refer to a target entity token $x^{(wiki)}$ given a discourse pre-context $X^{(pre)} = \lbrace x^{(pre)}_{1}, x^{(pre)}_{2}, ..., x^{(pre)}_{m} \rbrace$ and pos-context $X^{(pos)} = \lbrace x^{(pos)}_{1}, x^{(pos)}_{2}, ..., x^{(pos)}_{l} \rbrace$ with $m$ and $l$ tokens, respectively. The model is implemented as a multi-encoder, attention-decoder network with bidirectional \cite{schuster1997bidirectional} \ac{LSTM} \cite{hochreiter1997long} sharing the same input word-embedding matrix $V$, as explained further.

\subsection{Context encoders}

Our model starts by encoding the pre- and pos-contexts with two separate bidirectional \ac{LSTM} encoders \cite{schuster1997bidirectional,hochreiter1997long}. These modules learn feature representations of the text surrounding the target entity $x^{(wiki)}$, which are used for the referring expression generation. The pre-context $X^{(pre)} = \lbrace x^{(pre)}_{1}, x^{(pre)}_{2}, ..., x^{(pre)}_{m} \rbrace$ is represented by forward and backward  hidden-state vectors $(\overrightarrow{h}^{(pre)}_1, \cdots, \overrightarrow{h}^{(pre)}_m)$ and $(\overleftarrow{h}^{(pre)}_1, \cdots, \overleftarrow{h}^{(pre)}_m)$. The final annotation vector for each encoding timestep $t$ is obtained by the concatenation of the forward and backward representations  $h^{(pre)}_t = [\overrightarrow{h}^{(pre)}_t, \overleftarrow{h}^{(pre)}_t]$. The same process is repeated for the pos-context resulting in representations $(\overrightarrow{h}^{(pos)}_1, \cdots, \overrightarrow{h}^{(pos)}_l)$ and $(\overleftarrow{h}^{(pos)}_1, \cdots, \overleftarrow{h}^{(pos)}_l)$ and annotation vectors $h^{(pos)}_t = [\overrightarrow{h}^{(pos)}_t, \overleftarrow{h}^{(pos)}_t]$. Finally, the encoding of target entity $x^{(wiki)}$ is simply its entry in the shared input word-embedding matrix $V_{wiki}$. 

\subsection{Decoder}

The referring expression generation module is an \ac{LSTM} decoder implemented in 3 different versions: \texttt{Seq2Seq}, \texttt{CAtt} and \texttt{HierAtt}. 
All decoders at each timestep $i$ of the generation process 
take as input features their previous state $s_{i-1}$, 
the target entity-embedding $V_{wiki}$, the embedding of the previous
word of the referring expression $V_{y_{i-1}}$ and finally 
the summary vector of the pre- and pos-contexts $c_i$. The difference between the decoder variations is the method to compute $c_i$.

\paragraph{\texttt{Seq2Seq}} models the context vector $c_i$ at each timestep $i$ concatenating the pre- and pos-context annotation vectors averaged over time:

\begin{equation}
\footnotesize{
\hat{h}^{(pre)} = \frac{1}{N}\sum^N_i h^{(pre)}_i \\
}
\end{equation}
\begin{equation}
\footnotesize{
\hat{h}^{(pos)} = \frac{1}{N}\sum^N_i h^{(pos)}_i
}
\end{equation}
\begin{equation}
\footnotesize{
c_i = [\hat{h}^{(pre)}, \hat{h}^{(pos)}]
}
\end{equation}

\paragraph{\texttt{CAtt}} is an \ac{LSTM} decoder augmented with an attention mechanism \cite{bahdanau2014neural} over the pre- and pos-context encodings, which is used to compute $c_i$ at each timestep. We compute energies $e^{(pre)}_{ij}$ and $e^{(pos)}_{ij}$ between encoder states $h^{(pre)}_i$ and $h^{(post)}_i$  and decoder state $s_{i-1}$. These scores are normalized through the application of the softmax function to obtain the final attention probability $\alpha^{(pre)}_{ij}$ and $\alpha^{(post)}_{ij}$. Equations \ref{eq:energy} and \ref{eq:align} summarize the process with $k$ ranging over the two encoders ($k \in [pre, pos]$), being the projection matrices $W^{(k)}_a$ and $U^{(k)}_a$ and attention vectors $v^{(k)}_a$ trained parameters.

\begin{equation}
\footnotesize{
e^{(k)}_{ij} = v^{(k)T}_a \text{tanh}(W^{(k)}_a  s_{i-1}  + U^{(k)}_{a}  h^{(k)}_j ) 
}
\label{eq:energy}
\end{equation}

\begin{equation}
\footnotesize{
\alpha^{(k)}_{ij} = \frac{\text{exp}(e^{(k)}_{ij})}{\sum_{n=1}^{N} \text{exp}(e^{(k)}_{in})}
}
\label{eq:align}
\end{equation}

In general, the attention probability $\alpha_{ij}^{(k)}$ determines the amount of contribution of the $j$th token of $k$-context in the generation of the $i$th token of the referring expression. In each decoding step $i$, a final summary-vector for each context $c^{(k)}_i$ is computed by summing the encoder states $h^{(k)}_j$ weighted by the attention probabilities $\alpha^{(k)}_{i}$:

\begin{equation}
\footnotesize{
c^{(k)}_i = \sum_{j=1}^{N} \alpha^{(k)}_{ij} h^{(k)}_j
}
\end{equation}

To combine $c^{(pre)}_i$ and $c^{(pos)}_i$ into a single representation, 
this model simply concatenate the pre- and pos-context summary vectors $c_i = [c^{(pre)}_i, c^{(pos)}_i]$. 

\paragraph{\texttt{HierAtt}} implements a second attention mechanism inspired by \citet{libovicky2017attention} in order to generate attention weights for the pre- and pos-context summary-vectors $c^{(pre)}_i$ and $c^{(pos)}_i$ instead of concatenate them. Equations \ref{eq:hier1}, \ref{eq:hier2} and \ref{eq:hier3} depict the process, being the projection matrices $W^{(k)}_b$ and $U^{(k)}_b$ as well as attention vectors $v^{(k)}_b$ trained parameters ($k \in [pre, pos]$).

\begin{equation}
\footnotesize{
e^{(k)}_{i} = v^{(k)T}_b \text{tanh}(W^{(k)}_b  s_{i-1}  + U^{(k)}_b  c^{(k)}_i )
}
\label{eq:hier1}
\end{equation}

\begin{equation}
\footnotesize{
\beta^{(k)}_i = \frac{ \text{exp}(e^{(k)}_{i}) }{\sum_{n}^{} \text{exp}(e^{(n)}_i)}
\label{eq:hier2}
}
\end{equation}

\begin{equation}
\footnotesize{
c_i = \sum_{k} \beta^{(k)}_{i} U^{(k)}_b c^{(k)}_i
}
\label{eq:hier3}
\end{equation}

\paragraph{Decoding}

Given the summary-vector $c_i$, the embedding of the previous referring expression token $V_{y_{i-1}}$, the previous decoder state $s_{i-1}$ and the entity-embedding $V_{wiki}$, the decoders predict their next state which later is used to compute a probability distribution over the tokens in the output vocabulary for the next timestep as Equations \ref{eq:decoding} and \ref{eq:softmax} show.

\begin{equation}
\footnotesize{
s_i = \Phi_\text{dec}(s_{i-1}, [c_i, V_{y_{i-1}}, V_{wiki}])
}
\label{eq:decoding}
\end{equation}

\begin{align}
  \begin{split}
  p(y_{i}|y_{<i}, X^{(pre)}, & x^{(wiki)},  X^{(pos)}) = \\
  & \text{softmax}(W_c s_i + b)
  \end{split}
\label{eq:softmax}
\end{align}

In Equation \ref{eq:decoding}, $s_0$ and $c_0$ are zero-initialized vectors. In order to find the referring expression $y$ that maximizes the likelihood in Equation \ref{eq:softmax}, we apply a beam search with length normalization with $\alpha = 0.6$ \cite{wu2016}:

\begin{equation}
\footnotesize{
lp(y) = \frac{(5+|y|)^{\alpha}}{(5+1)^{\alpha}}
\label{eq:lengthnorm}
}
\end{equation}

The decoder is trained to minimize the negative log likelihood of the next token
in the target referring expression:
\begin{equation}
\footnotesize{
J(\theta) = - \sum_{i} \text{log p}(y_i|y_{<i}, X^{(pre)}, x^{(wiki)}, X^{(pos)})
}
\end{equation}

\section{Models for Comparison}

We compared the performance of NeuralREG against two baselines: \textit{OnlyNames} and a model based on the choice of referential form method of \citet{ferreira2016b}, dubbed \textit{Ferreira}.
\paragraph{OnlyNames} is motivated by the similarity among the Wikipedia ID of an element and a proper name reference to it. This method refers to each entity by their Wikipedia ID, replacing each underscore in the ID for whitespaces (e.g., \textit{Appleton\_International\_Airport} to \textit{``Appleton International Airport''}).

\paragraph{Ferreira} works by first choosing whether a reference should be a proper name, pronoun, description or demonstrative. The choice is made by a Naive Bayes method as Equation \ref{eq:bayes} depicts.

\begin{equation}
\footnotesize{
P(f \mid X) \propto \frac{P(f) \prod\limits_{x \in X} P(x \mid f)}{\sum\limits_{f' \in F} P(f') \prod\limits_{x \in X} P(x \mid f')} 
\label{eq:bayes}
}
\end{equation}

The method calculates the likelihood of each referential form $f$ given a set of features $X$, consisting of grammatical position and information status (new or given in the text and sentence). Once the choice of referential form is made, the most frequent variant is chosen in the training corpus given the referent, syntactic position and information status. In case a referring expression for a wiki target is not found in this way, a back-off method is applied by removing one factor at a time in the following order: sentence information status, text information status and grammatical position. Finally, if a referring expression is not found in the training set for a given entity, the same method as \textit{OnlyNames} is used. Regarding the features, syntactic position distinguishes whether a reference is the subject, object or subject determiner (genitive) in a sentence. Text and sentence information statuses mark whether a reference is a initial or a subsequent mention to an entity in the text and the sentence, respectively. All features were extracted automatically from the texts using the sentence tokenizer and dependency parser of Stanford CoreNLP \cite{manning2014}. 

\section{Automatic evaluation}

\paragraph{Data} We evaluated our models on the training, development and test referring expression sets described in Section \ref{sec:reg_corpus}.

\paragraph{Metrics} We compared the referring expressions produced by the evaluated models with the gold-standards ones using accuracy and String Edit Distance \cite{levenshtein1966}. Since pronouns are highlighted as the most likely referential form to be used when a referent is salient in the discourse, as argued in the introduction, we also computed pronoun accuracy, precision, recall and F1-score in order to evaluate the performance of the models for capturing discourse salience. Finally, we lexicalized the original templates with the referring expressions produced by the models and compared them with the original texts in the corpus using accuracy and BLEU score \cite{papineni2002} as a measure of fluency. Since our model does not handle referring expressions for constants (dates and numbers), we just copied their source version into the template.

Post-hoc McNemar's and Wilcoxon signed ranked tests adjusted by the Bonferroni method were used to test the statistical significance of the models in terms of accuracy and string edit distance, respectively. To test the statistical significance of the BLEU scores of the models, we used a bootstrap resampling together with an approximate randomization method \cite{clarketal2011}\footnote{\url{https://github.com/jhclark/multeval}}. 

\paragraph{Settings} NeuralREG was implemented using Dynet \cite{neubig2017}. Source and target word embeddings were 300D each and trained jointly with the model, whereas hidden units were 512D for each direction, totaling 1024D in the bidirection layers. All non-recurrent matrices were initialized following the method of \citet{glorot2011}. Models were trained using stochastic gradient descent with Adadelta \cite{zeiler2012} and mini-batches of size 40. We ran each model for 60 epochs, applying early stopping for model selection based on accuracy on the development set with patience of 20 epochs. For each decoding version (\texttt{Seq2Seq}, \texttt{CAtt} and \texttt{HierAtt}), we searched for the best combination of drop-out probability of 0.2 or 0.3 in both the encoding and decoding layers, using beam search with a size of 
1 or 5 with predictions up to 30 tokens or until 2 ending tokens were predicted (\textit{EOS}). The results described in the next section were obtained on the test set by the NeuralREG version with the highest accuracy on the development 
set over the epochs.

\begin{table*}
\footnotesize{
\centering
	\begin{tabular}{l l l | l l l l | l l}
    \midrule
    & \multicolumn{2}{ c |}{\textbf{All References}} & \multicolumn{4}{ c |}{\textbf{Pronouns}} & \multicolumn{2}{ c }{\textbf{Text}}  \\
	& Acc. & SED & Acc. & Prec. & Rec. & F-Score & Acc. & BLEU  \\
    \midrule
    \textit{OnlyNames}        & 0.53$^{D}$ & 4.05$^{D}$ & - & - & - & - & 0.15$^{D}$ & 69.03$^{D}$  \\
    \textit{Ferreira}          & 0.61$^{C}$ & 3.18$^{C}$ & 0.43$^{B}$ & 0.57 & 0.54 & 0.55 & 0.19$^{C}$ & 72.78$^{C}$  \\
    \midrule
    NeuralREG+\texttt{Seq2Seq} & 0.74$^{A,B}$ & 2.32$^{A,B}$ & 0.75$^{A}$ & 0.77 & 0.78 & 0.78 & 0.28$^{B}$ & 79.27$^{A,B}$  \\
    NeuralREG+\texttt{CAtt}    & 0.74$^{A}$ & 2.25$^{A}$ & 0.75$^{A}$ & 0.73 & 0.78 & 0.75 & 0.30$^{A}$  & 79.39$^{A}$  \\
    NeuralREG+\texttt{HierAtt} & 0.73$^{B}$ & 2.36$^{B}$ & 0.73$^{A}$ & 0.74 & 0.77 & 0.75 & 0.28$^{A,B}$ & 79.01$^{B}$  \\
    \bottomrule
	\end{tabular}
\caption{(1) Accuracy (Acc.) and String Edit Distance (SED) results in the prediction of all referring expressions; (2) Accuracy (Acc.), Precision (Prec.), Recall (Rec.) and F-Score results in the prediction of pronominal forms; and (3) Accuracy (Acc.) and BLEU score results of the texts with the generated referring expressions. Rankings were determined by statistical significance.}
\label{table:results}
}
\end{table*}

\paragraph{Results} Table \ref{table:results} summarizes the results for all models on all metrics on the test set and Table \ref{table:example} depicts a text example lexicalized by each model. The first thing to note in the results of the first table is that the baselines in the top two rows performed quite strong on this task, generating more than half of the referring expressions exactly as in the gold-standard. The method based on \citet{ferreira2016b} performed statistically better than {\it OnlyNames} on all metrics due to its capability, albeit to a limited extent, to predict pronominal references (which {\it OnlyNames}\/ obviously cannot). 

We reported results on the test set for NeuralREG+\texttt{Seq2Seq} and NeuralREG+\texttt{CAtt} using dropout probability 0.3 and beam size 5, and  NeuralREG+\texttt{HierAtt} with dropout probability of 0.3 and beam size of 1 selected based on the highest accuracy on the development set. Importantly, the three NeuralREG variant models statistically outperformed the two baseline systems. They achieved BLEU scores, text and referential accuracies as well as string edit distances in the range of 79.01-79.39, 28\%-30\%, 73\%-74\% and 2.25-2.36, respectively. This means that NeuralREG predicted 3 out of 4 references completely correct, whereas the incorrect ones needed an average of 2 post-edition operations in character level to be equal to the gold-standard. When considering the texts lexicalized with the referring expressions produced by NeuralREG, at least 28\% of them are similar to the original texts. Especially noteworthy was the score on pronoun accuracy, indicating that the model was well capable of predicting when to generate a pronominal reference in our dataset.

The results for the different decoding methods for NeuralREG were similar, with the NeuralREG+\texttt{CAtt} performing slightly better in terms of the BLEU score, text accuracy and String Edit Distance. The more complex NeuralREG+\texttt{HierAtt} yielded the lowest results, even though the differences with the other two models were small and not even statistically significant in many of the cases.

\begin{table*}
\footnotesize{
\begin{center}
\begin{tabular}{l L}
\hline
\textbf{Model} & \textbf{Text} \\
\hline
\textit{OnlyNames}      & \textbf{alan shepard} was born in \textbf{new hampshire} on \textbf{1923-11-18} . before \textbf{alan shepard} death in \textbf{california} \textbf{alan shepard} had been awarded \textbf{distinguished service medal (united states navy)} an award higher than \textbf{department of commerce gold medal} . \\
\hline
\textit{Ferreira}   & \textbf{alan shepard} was born in \textbf{new hampshire} on \textbf{1923-11-18} . before \textbf{alan shepard} death in \textbf{california} \textbf{him} had been awarded \textbf{distinguished service medal} an award higher than \textbf{department of commerce gold medal} .
 \\
\hline
\texttt{Seq2Seq}   & \textbf{alan shepard} was born in \textbf{new hampshire} on \textbf{1923-11-18} . before \textbf{his} death in \textbf{california} \textbf{him} had been awarded \textbf{the distinguished service medal by the united states navy} an award higher than \textbf{the department of commerce gold medal} . \\
\hline
\texttt{CAtt}   & \textbf{alan shepard} was born in \textbf{new hampshire} on \textbf{1923-11-18} . before \textbf{his} death in \textbf{california} \textbf{he} had been awarded \textbf{the distinguished service medal by the us navy} an award higher than \textbf{the department of commerce gold medal} . \\
\hline
\texttt{HierAtt}   & \textbf{alan shephard} was born in \textbf{new hampshire} on \textbf{1923-11-18} . before \textbf{his} death in \textbf{california} \textbf{he} had been awarded \textbf{the distinguished service medal} an award higher than \textbf{the department of commerce gold medal} . \\
\hline
Original    & \textbf{alan shepard} was born in \textbf{new hampshire} on \textbf{18 november 1923} . before \textbf{his} death in \textbf{california} \textbf{he} had been awarded \textbf{the distinguished service medal by the us navy} an award higher than \textbf{the department of commerce gold medal} . \\
\hline
\end{tabular}
\caption{Example of text with references lexicalized by each model.}
\label{table:example}
\end{center}
}
\end{table*}

\section{Human Evaluation}
Complementary to the automatic evaluation, we performed an evaluation with human judges, comparing the quality judgments of the original texts to the versions generated by our various models.

\paragraph{Material} We quasi-randomly selected 24 instances from the delexicalized version of the WebNLG corpus related to the test part of the referring expression collection. For each of the selected instances,  we took into account its source triple set and its 6 target texts: one original (randomly chosen) and its versions with the referring expressions generated by each of the 5 models introduced in this study (two baselines, three neural models). Instances were chosen following 2 criteria: the number of triples in the source set (ranging from 2 to 7) and the differences between the target texts. 

For each size group, we randomly selected 4 instances (of varying degrees of variation between the generated texts) giving rise to 144 trials ($=$ 6 triple set sizes $*$ 4 instances $*$ 6 text versions), each consisting of a set of triples and a target text describing it with the lexicalized referring expressions highlighted in yellow.

\paragraph{Method} The experiment had a latin-square design, distributing the 144 trials over 6 different lists such that each participant rated 24 trials, one for each of the 24 corpus instances, making sure that participants saw equal numbers of triple set sizes and generated versions. Once introduced to a trial, the participants were asked to rate the fluency (``does the text flow in a natural, easy to read manner?''), grammaticality (``is the text grammatical (no spelling or grammatical errors)?'') and clarity (``does the text clearly express the data?") of each target text on a 7-Likert scale, focussing on the highlighted referring expressions. The experiment is available on the website of the author\footnote{\url{https://ilk.uvt.nl/~tcastrof/acl2018/evaluation/}}. 

\paragraph{Participants} We recruited 60 participants, 10 per list, via Mechanical Turk. Their average age was 36 years and 27 of them were females. The majority declared themselves native speakers of English (44), while 14 and 2 self-reported as fluent or having a basic proficiency, respectively.
 
 \begin{table}
\footnotesize{
\centering
	\begin{tabular}{l l l l}
    \midrule
	& Fluency & Grammar & Clarity \\
    \midrule
    \textit{OnlyNames}        & 4.74$^{C}$ & 4.68$^{B}$ & 4.90$^{B}$ \\
    \textit{Ferreira}          & 4.74$^{C}$ & 4.58$^{B}$ & 4.93$^{B}$ \\
    \midrule
    NeuralREG+\texttt{Seq2Seq} & 4.95$^{B,C}$ & 4.82$^{A,B}$ & 4.97$^{B}$ \\
    NeuralREG+\texttt{CAtt}    & 5.23$^{A,B}$ & 4.95$^{A,B}$ & 5.26$^{A,B}$ \\
    NeuralREG+\texttt{HierAtt} & 5.07$^{B,C}$ & 4.90$^{A,B}$ & 5.13$^{A,B}$ \\
    \midrule
    \textit{Original}          & 5.41$^{A}$ & 5.17$^{A}$ & 5.42$^{A}$ \\
    \bottomrule
	\end{tabular}
\caption{Fluency, Grammaticality and Clarity results obtained in the human evaluation. Rankings were determined by statistical significance.}
\label{table:human}
}
\end{table}
 
 \paragraph{Results}

Table \ref{table:human} summarizes the results. Inspection of the Table reveals a clear pattern: all three neural models scored higher than the baselines on all metrics, with especially NeuralREG+\texttt{CAtt} approaching the ratings for the original sentences, although -- again -- differences between the neural models were small. Concerning the size of the triple sets, we did not find any clear pattern.

To test the statistical significance of the pairwise comparisons, we used the Wilcoxon signed-rank test corrected for multiple comparisons using the Bonferroni method. Different from the automatic evaluation, the results of both baselines were not statistically significant for the three metrics. In comparison with the neural models, NeuralREG+\texttt{CAtt} significantly outperformed the baselines in terms of fluency, whereas the other comparisons between baselines and neural models were not statistically significant. The results for the 3 different decoding methods of NeuralREG also did not reveal a significant difference. Finally, the original texts were rated significantly higher than both baselines in terms of the three metrics, also than NeuralREG+\texttt{Seq2Seq} and NeuralREG+\texttt{HierAtt} in  terms of fluency, and than NeuralREG+\texttt{Seq2Seq} in terms of clarity.

\section{Discussion}

This study introduced NeuralREG, an end-to-end approach based on neural networks which tackles the full Referring Expression Generation process. It generates referring expressions for discourse entities by simultaneously selecting form and content without any need of feature extraction techniques. The model was implemented using an encoder-decoder approach where a target referent and its surrounding linguistic contexts were first encoded and combined into a single vector representation which subsequently was decoded into a referring expression to the target, suitable for the specific discourse context. In an automatic evaluation on a collection of 78,901 referring expressions to 1,501 Wikipedia entities, the different versions of the model all yielded better results than the two (competitive) baselines. Later in a complementary human evaluation, the texts with referring expressions generated by a variant of our novel model were considered statistically more fluent than the texts lexicalized by the two baselines.

\paragraph{Data} The collection of referring expressions used in our experiments was extracted from a novel, delexicalized and publicly available version of the WebNLG corpus \cite{claire2017,claire2017b}, where the discourse entities were replaced with general tags for decreasing the data sparsity. Besides the \ac{REG} task, these data can be useful for many other tasks related to, for instance, the \ac{NLG} process \cite{reiter2000,gatt2017} and Wikification~\cite{moussallem2017}.

\paragraph{Baselines} We introduced two strong baselines which generated roughly half of the referring expressions identical to the gold standard in an automatic evaluation. These baselines performed relatively well because they frequently generated full names, which occur often for our wikified references. However, they performed poorly when it came to pronominalization, which is an important ingredient for fluent, coherent text. \textit{OnlyNames}, as the name already reveals, does not manage to generate any pronouns. However, the approach of \citet{ferreira2016b} also did not perform well in the generation of pronouns, revealing a poor capacity to detect highly salient entities in a text.

\paragraph{NeuralREG} was implemented with 3 different decoding architectures: \texttt{Seq2Seq}, \texttt{CAtt} and \texttt{HierAtt}. Although all the versions performed relatively similar, the concatenative-attention (\texttt{CAtt}) version generated the closest referring expressions from the gold-standard ones and presented the highest textual accuracy in the automatic evaluation. The texts lexicalized by this variant were also considered statistically more fluent than the ones generated by the two proposed baselines in the human evaluation. 

Surprisingly, the most complex variant (\texttt{HierAtt}) with a hierarchical-attention mechanism gave lower results than \texttt{CAtt}, producing lexicalized texts which were rated as less fluent  than the original ones and not significantly more fluent from the ones generated by the baselines. This result appears to be not consistent with the findings of \citet{libovicky2017attention}, who reported better results on multi-modal machine translation with hierarchical-attention as opposed to the flat variants \cite{specia2016shared}. 

Finally, our NeuralREG variant with the lowest results were our `vanilla'  sequence-to-sequence (\texttt{Seq2Seq}),  whose the lexicalized texts were significantly less fluent and clear than the original ones. This shows the importance of the attention mechanism in the decoding step of NeuralREG in order to generate fine-grained referring expressions in discourse.

\paragraph{Conclusion} We introduced a deep learning model for the generation of referring expressions in discourse texts. NeuralREG decides both on referential form and on referential content in an integrated, end-to-end approach, without using explicit features. Using a new delexicalized version of the WebNLG corpus (made publicly available), we showed that the neural model substantially improves over two strong baselines in terms of accuracy of the referring expressions and fluency of the lexicalized texts.



\section*{Acknowledgments}
This  work  has  been  supported  by  the  National Council of Scientific and Technological Development from Brazil (CNPq) under the grants 203065/2014-0 and 206971/2014-1.

\bibliography{naaclhlt2018}
\bibliographystyle{acl_natbib.bst}

\end{document}